\documentclass[journal,twoside,web]{ieeecolor}
\usepackage{generic}
\usepackage{cite}
\usepackage{amsmath,amssymb,amsfonts}
\usepackage{algorithmic}
\usepackage{graphicx}
\usepackage{textcomp}
\usepackage{multirow}
\usepackage{bbding}
\usepackage{flushend}
\usepackage{threeparttable}
\usepackage{url}
\usepackage{makecell}

\def\BibTeX{{\rm B\kern-.05em{\sc i\kern-.025em b}\kern-.08em
		T\kern-.1667em\lower.7ex\hbox{E}\kern-.125emX}}
\begin{document}
	\title{BATFormer: Towards Boundary-Aware Lightweight Transformer for Efficient Medical Image Segmentation}
	\author{Xian Lin, Li Yu, \IEEEmembership{Senior Member, IEEE}, Kwang-Ting Cheng, \IEEEmembership{Fellow, IEEE}, and Zengqiang Yan
		\thanks{Corresponding author: Zengqiang Yan.}
		\thanks{Xian Lin, Li Yu, and Zengqiang Yan are with the School of Electronic Information and Communications, Huazhong University of Science and Technology, Wuhan 430074, China (e-mail: xianlin@hust.edu.cn; hustlyu@hust.edu.cn; z\_yan@hust.edu.cn).}
		\thanks{Kwang-Ting Cheng is with the School of Engineering, Hong Kong University of Science and Technology, Kowloon, Hong Kong (e-mail: timcheng@ust.hk).}
	}
	
	\maketitle

	\begin{abstract}
			Objective: Transformers, born to remedy the inadequate receptive fields of CNNs, have drawn explosive attention recently. However, the daunting computational complexity of global representation learning, together with rigid window partitioning, hinders their deployment in medical image segmentation. This work aims to address the above two issues in transformers for better medical image segmentation. Methods: We propose a boundary-aware lightweight transformer (BATFormer) that can build cross-scale global interaction with lower computational complexity and generate windows flexibly under the guidance of entropy. Specifically, to fully explore the benefits of transformers in long-range dependency establishment, a cross-scale global transformer (CGT) module is introduced to jointly utilize multiple small-scale feature maps for richer global features with lower computational complexity. Given the importance of shape modeling in medical image segmentation, a boundary-aware local transformer (BLT) module is constructed. Different from rigid window partitioning in vanilla transformers which would produce boundary distortion, BLT adopts an adaptive window partitioning scheme under the guidance of entropy for both computational complexity reduction and shape preservation. Results: BATFormer achieves the best performance in Dice of 92.84$\%$, 91.97$\%$, 90.26$\%$, and 96.30$\%$ for the average, right ventricle, myocardium, and left ventricle respectively on the ACDC dataset and the best performance in Dice, IoU, and ACC of 90.76$\%$, 84.64$\%$, and 96.76$\%$ respectively on the ISIC 2018 dataset. More importantly, BATFormer requires the least amount of model parameters and the lowest computational complexity compared to the state-of-the-art approaches. Conclusion and Significance: Our results demonstrate the necessity of developing customized transformers for efficient and better medical image segmentation. We believe the design of BATFormer is inspiring and extendable to other applications/frameworks. The source code is publicly available at \url{https://github.com/xianlin7/BATFormer}.	
	\end{abstract}
	
	\section{Introduction}
	\label{sec:introduction}
	
	\IEEEPARstart{M}{edical} image segmentation is a vital technique for computer-assisted diagnosis. Through progressively merging multi-level features with skip connections, U-shaped convolutional neural networks (CNN) have dominated the field of medical image segmentation with superior performance \cite{r1, r2, r3, r4, r5}. However, these brilliant CNN-based architectures still suffer from limited receptive fields.
	
	Transformers \cite{r5_0, r22, r5_2, r5_3, r5_4}, born to capture long-term interaction, have recently attracted extensive attention in medical image segmentation \cite{r5_5, r5_6, r5_7, r5_8, r5_9}. Chen \textit{et al.} \cite{r6} first explored the potential of transformers in medical image segmentation by introducing transformers to the last encoder layer of the vanilla U-Net framework. Then, a series of frameworks have been proposed focusing on combining transformer- and CNN-blocks in encoders for better feature modeling \cite{r7, r8}. However, directly coupling vanilla transformers with the encoder to build global dependency usually results in daunting computational complexity, making it lumber-some for representation learning on high-resolution feature maps.
	To make transformers more computationally efficient, CoTr \cite{r9} introduced deformable self-attention to make full use of just a small set of key points.
	SwinUnet \cite{r10} divided feature maps into a series of shifted windows and performed dependency establishment within each window separately. UTNet \cite{r11} proposed an efficient self-attention mechanism to reduce computational complexity by projecting $K$ and $V$ into low dimensions. 
	Similarly, GT U-Net \cite{r12} introduced a grouping structure and a bottleneck structure for lightweight deployment. 
	While effective in complexity reduction, the above frameworks suffer from vital local feature loss and insufficient robustness for multi-scale object segmentation due to rigid patch/window partitioning.
	
	Recently, several approaches have been proposed to preserve the locality of transformer-based architectures. DS-TransUNet \cite{r13} adopted dual-scale Swin transformer blocks to extract features under different window divisions, based on which to restore the window internal structure destroyed by single-scale window division. 
	TransAttUnet \cite{r14} introduced multi-scale skip connections among decoder layers to minimize information loss. 
	MISSFormer \cite{r15} redesigned the feed-forward network with depth-wise convolution to supply locality. 
	MT-UNet \cite{r16} proposed a local-global Gaussian-weighted self-attention by alternating self-attention with local and global receptive fields. 
	Despite the effectiveness in increasing the locality of transformers, the partitioning schemes of these approaches still are rigid, struggling to segment objects with irregular shapes and varying sizes. 
	
	\begin{figure}[!t]
		\centering
		\includegraphics[width=1\columnwidth]{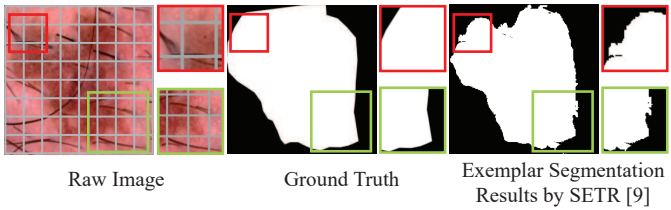}
		\caption{Rigid window/patch partitioning of transformers resulting in boundary distortion. The gray grids in the raw image describe the window/patch partitioning results of vanilla transformers, which may destroy important boundary details. Boxes in red and green zoom in the corresponding regions/patches.}
		\label{fig1}
	\end{figure}
	
	In this paper, we propose a boundary-aware lightweight transformer, named BATFormer, for efficient medical image segmentation. Specifically, a cross-scale global transformer (CGT) module is constructed to localize the rough areas of objects by capturing global dependency across multiple small-scale features. 
	In this way, richer long-range dependency is established with lower computational costs and the overall features of large objects can be well preserved. To overcome the boundary distortion problem introduced by rigid window partitioning in vanilla transformers, a boundary-aware local transformer (BLT) module is deployed to adaptively generate windows around object boundaries and perform self-attention within each window. In this way, BLT could better localize boundaries for shape preservation which is crucial for diagnosis in practice.
	Qualitative and quantitative comparison results with the state-of-the-art methods on two public datasets, including the ISIC 2018 dataset \cite{r17,r18} and the ACDC dataset \cite{r19}, demonstrate the effectiveness of BATFormer for both 2D and 3D medical image segmentation.
	More importantly, BATFormer incurs much lower computational complexity, demonstrating the necessity of developing lightweight transformers for medical image segmentation. The contributions are summarized as follows:
	\begin{itemize}
		\item BATFormer, a boundary-aware lightweight transformer for medical image segmentation, optimizing transformers to better handle practical scenarios consisting of various shapes of objects.
		\item A cross-scale global transformer (CGT) module builds richer global-range dependency with much lower computational complexity for multi-scale object segmentation.
		\item A boundary-aware local transformer (BLT) module captures mid-range dependency for hard shape preservation. To our best knowledge, this is the first attempt to study dynamic and adaptive window partitioning for medical image segmentation.
		\item Superior performance against the state-of-the-art CNN-based and transformer-based/-hybrid approaches on publicly-available datasets covering 2D and 3D various sizes and shapes of segmentation objects.
	\end{itemize}
	
	The rest of this paper is organized as follows. Section \ref{sec:problem} analyzes the deployment challenges of transformers for medical image segmentation and Section \ref{sec:method} describes the proposed BATFormer in detail. We present a thorough evaluation of BATFormer against the state-of-the-art CNN-based and transformer-based-hybrid methods in Section \ref{sec:evaluation} and ablation studies in Section \ref{sec:discussion}. Section \ref{sec:conclusion} concludes this paper.
	
	\section{Problem Analysis}
	\label{sec:problem}
	
	In medical image segmentation, objects, \textit{e.g.} organs, lesions, tissues, \textit{etc.}, can have varying sizes and irregular shapes, segmenting which often suffers from under-segmentation and poor boundary detection for both CNN-based and transformer-based approaches, due to the following reasons:
	\begin{enumerate}
		\item \textbf{Inadequate receptive fields of CNN}: Medical image segmentation relies on both global and local features for semantic capturing and detail localization. Unfortunately, the locality nature of CNN makes it struggle to extract rich global features to deal with local contextual similarity, resulting in under-segmentation.
		\item \textbf{Inefficient architecture design of vanilla transformers}: Transformer is supposed to complement CNN for global representation learning, which builds pairwise long-range dependency between each pair of tokens/patches through self-attention.
		However, vanilla transformers construct global receptive fields at the cost of cumbersome computational complexity, limiting its deployment to extracting richer semantic information (\textit{i.e.} cross-/multi-scale global semantic features). In addition, when transformers are deployed on high-resolution feature maps, they usually adopt window-wise self-attention for mid-range dependency building. Unfortunately, such window partitioning, where each image is simply divided into windows through evenly tiled grids, is rigid which can be fatal for accurate boundary detection. As illustrated in Fig.~\ref{fig1}, detailed information around split lines would be seriously destroyed, which negatively affects boundary localization.
	\end{enumerate}

	\begin{figure*}[!t]
		\centering
		\includegraphics[width=1\textwidth]{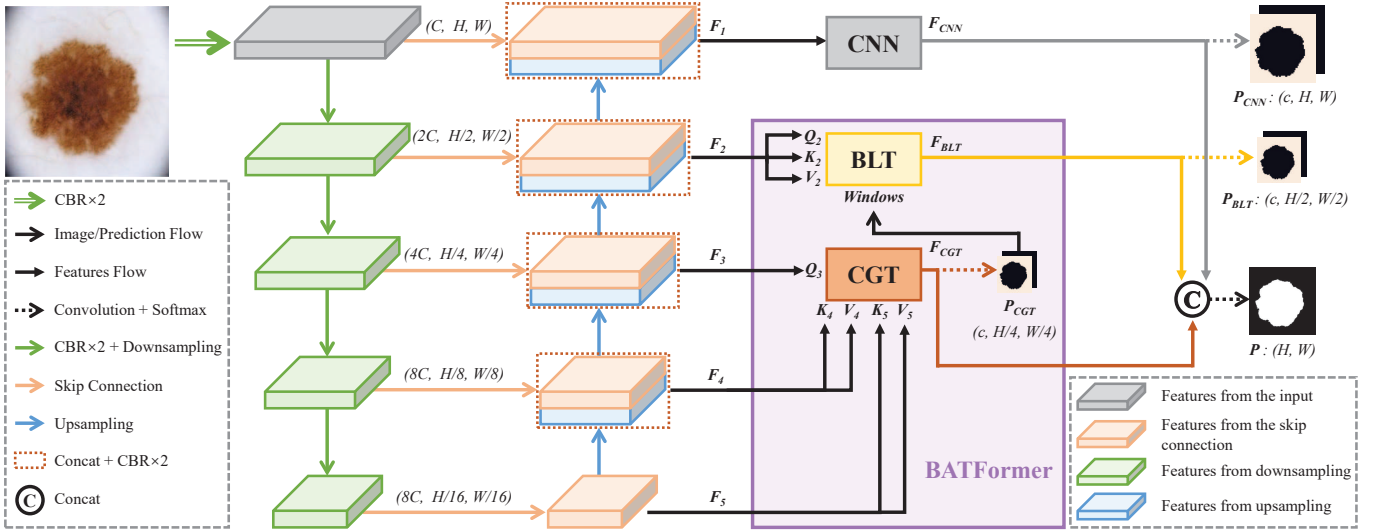}
		\caption{Overview of BATFormer. CGT and BLT represent the cross-scale global transformer module and the boundary-aware local transformer module respectively. Functional composition CBR denotes convolution, batch normalization, and Relu.}
		\label{fig2}
	\end{figure*}

	To better complement CNN, a good transformer design should fulfill the following two requirements:
	\begin{enumerate}
		\item \textbf{Multi-scale long-range dependency establishment with acceptable computational costs}: Building and fusing multi-scale long-range dependency is helpful to enforce transformers to focus more on global representation learning. However, directly applying vanilla transformer blocks to multi-scale feature maps would result in high computational costs. How to balance computational complexity (\textit{i.e.} lightweight) and multi-scale long-range dependency establishment (\textit{i.e.} effectiveness) is the main challenge.
		\item \textbf{Content-aware partitioning}: Instead of uniformly and evenly patch/window generation, window/patch partitioning should be carefully conducted to preserve valuable contextual information, especially for the boundary regions. To our best knowledge, flexible window partitioning in transformers has been rarely explored.
	\end{enumerate}
	To this end, we propose a novel transformer-hybrid architecture BATFormer by re-designing transformers according to the above characteristics for efficient medical image segmentation.
	
	\section{Method}
	\label{sec:method}
	
	\subsection{Overview}
	
	The overall architecture of BATFormer is depicted in Fig.~\ref{fig2}, consisting of a lightweight U-shaped backbone and two major components to address each of the above issues:
	
	\begin{enumerate}
		\item \textbf{Cross-Scale Global Transformer (CGT)}:
		On the one hand, building global transformers on large-scale representations (\textit{e.g.} $F_1$ and $F_2$) is highly expensive as the computational complexity is proportional to the square of the length of the input sequences. On the other hand, global dependency is mainly required for constructing deep semantic features. Therefore, only multiple small-scale (\textit{i.e.} high-level) feature maps (\textit{i.e.}, $F_3$, $F_4$, and $F_5$) are fed to CGT. Through cross-scale self-attention, CGT manages to extract richer semantic features for global representation learning with lower computational complexity, which in turn complements the CNN module for joint local and global feature extraction and fusion.
		\item \textbf{Boundary-Aware Local Transformer (BLT)}: To overcome the boundary distortion issue, a flexible window partitioning scheme is proposed to localize boundary-related windows based on entropy calculation. Following this, BLT can filter out non-boundary windows and better focus on exploring long-range features for boundary detection with lower computational complexity.
	\end{enumerate}
	In the following, we detail each component of BATFormer.
	
	\subsection{Lightweight U-shaped Backbone}
	The lightweight U-shaped backbone mainly consists of four downsampling blocks, four up-sampling blocks, and five skip connections. Specifically, each input image would be fed into four consecutive downsampling blocks after two $3\times3$ convolutions and then be decoded by four upsampling blocks. Each downsampling block consists of one $2\times2$ max pooling with stride 2 and two $3\times3$ convolutions while each upsampling block includes bilinear interpolation, concatenation, and two $3\times3$ convolutions. Downsampling and upsampling blocks of the same scales are connected by skip connections as U-Net~\cite{r1}. As depicted in Fig.~\ref{fig2}, the channel numbers of convolutions in the downsampling and upsampling blocks are set as ($2C$, $4C$, $8C$, $8C$), and ($4C$, $2C$, $C$, $C$) where $C$ is 16.
	
	\subsection{Cross-Scale Global Transformer}
	
	\begin{figure*}[!t]
		\centering
		\includegraphics[width=1\textwidth]{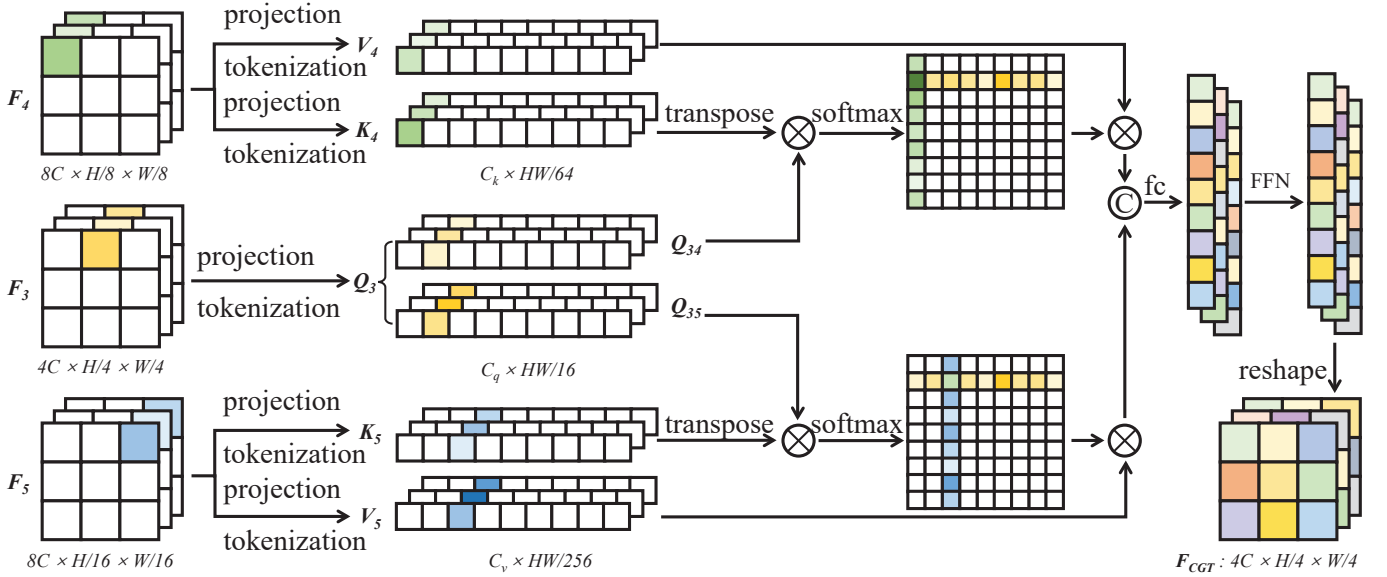}
		\caption{Overview of the cross-scale global transformer (CGT) module. The residual structures of self-attention and the feed-forward neural networks (FFN) are omitted for brevity.}
		\label{fig3}
	\end{figure*}

	CGT consists of two cross-scale attention modules and one feed-forward network (FFN) as illustrated in Fig.~\ref{fig3}, aiming to build global dependency among the three small-scale feature maps generated by the backbone, \textit{i.e.,} $F_3 \in \mathbb{R} ^ {4C \times \frac{H}{4} \times \frac{W}{4}}$, $F_4 \in \mathbb{R} ^ {8C \times \frac{H}{8} \times \frac{W}{8}}$, and $F_5 \in \mathbb{R} ^ {8C \times \frac{H}{16} \times \frac{W}{16}}$, where $C$ is the channel count of the first branch head and $(H, W)$ is the resolution of the input image. In the cross-scale attention module, $F_3$ with the maximum resolution is projected into two queries $Q_{3,4} \in \mathbb{R} ^ {\frac{HW}{16} \times d}$ and $Q_{3,5} \in \mathbb{R} ^ {\frac{HW}{16} \times d}$ where $d$ is the dimension of the transformer module, while $F_4$ and $F_5$ are projected into two key-value groups \{$K_4$, $V_4$\} $\in \mathbb{R} ^ {\frac{HW}{64} \times d}$ and \{$K_5$, $V_5$\} $\in \mathbb{R} ^ {\frac{HW}{256} \times d}$. Then, cross-scale attention is written as
	\begin{equation}
		\mathcal{F}^i_{ca}(Q_{3,i}, K_i, V_i) = softmax(\frac{Q_{3,i}K_i^T}{\sqrt{d}})V_i,
		\label{eq1}
	\end{equation}
	where $i$ is set as 4 and 5 for $F_4$ and $F_5$ respectively. In contrast to the vanilla self-attention which generates $Q$, $K$, and $V$ from the same input, the proposed cross-scale attention derives $K$ and $V$ from the other two smaller-scale feature maps. In this way, computational complexity can be reduced by a factor of $2^2$ or $2^4$ since the sequence lengths of $K$ and $V$ are shorter. More importantly, diverse features are included for dependency establishment, since different groups of $K$ and $V$ correspond to different scales of receptive fields and semantic information.
	
	Before feeding to FFN, the two sets of cross-scale attention are combined and refined by
	\begin{equation}
		F_{ca} = (\mathcal{F}^4_{ca,1} \textcircled{c} \cdot \cdot \cdot \textcircled{c} \mathcal{F}^4_{ca,g} \textcircled{c} \mathcal{F}^5_{ca,1} \cdot \cdot \cdot \textcircled{c} \mathcal{F}^5_{ca,g}) \cdot W_{ca} \textcircled{+} F_3,
		\label{eq2}
	\end{equation}
	where $g$ is the predefined number of transformer heads in CGT, $W_{ca} \in  \mathbb{R} ^ {2gd \times d}$ is a learnable projection matrix for combination, $\textcircled{c}$ is the concatenation operation, and $\textcircled{+}$ represents the residual connection via element-wise addition.
	
	FFN after cross-scale attention is further processed to obtain the final output of CGT according to
	\begin{equation}
		F_{CGT} = (max(0, F_{ca}W_{c1}+b_{c1}) \cdot W_{c2}+b_{c2}) \textcircled{+} F_{ca},
		\label{eq3}
	\end{equation}
	where $W_{c1} \in \mathbb{R} ^ {d \times 4d} $ and  $W_{c2} \in \mathbb{R} ^ {4d \times d} $ are learnable projection matrices, and $b_{c1}$, $b_{c2}$ $\in \mathbb{R}$ are the offsets.
	
	\subsection{Boundary-Aware Local Transformer}
	
	BLT contains three stages: dynamic boundary-aware window generation for adaptively boundary localization, boundary-around self-attention for boundary identification, and FFN for feature distillation. As analyzed in Section II, existing transformers adopt rigid window/patch partitioning, which severely destroys the vital details around boundaries. Therefore, the core of BLT is to perform local self-attention within boundary-aware windows, which are generated under the guidance of entropy calculation. Firstly, evenly and densely tiled windows over the feature map $F_2 \in \mathbb{R} ^ {2C \times \frac{H}{2} \times \frac{W}{2}}$ are collected as the initial window set $\{w\}$ containing all possible window positions. Before calculating the probability of each window being boundary-around, the entropy of each position $(m,n)$ in the probability map $P_{CGT} \in \mathbb{R} ^ {c \times \frac{H}{4} \times \frac{W}{4}}$ produced by CGT is calculated by
	\begin{equation}
		\mathcal{C}_p{(m, n)} = -\frac{1}{{\log_2}c}\sum_{l=1}^{c}P_{CGT}(l, m, n){\log_2}P_{CGT}(l, m, n),
		\label{eq4}
	\end{equation}
	where $c$ is the number of classes. Then, the score of each window is defined as:
	\begin{equation}
		\mathcal{C}_w{(x, y)} = \frac{1}{\lfloor \frac{h}{2} \rfloor \lfloor \frac{w}{2} \rfloor} \sum_{m=0}^{\lfloor \frac{h}{2} \rfloor}\sum_{n=0}^{\lfloor \frac{w}{2} \rfloor}	\mathcal{C}_p{(\lfloor \frac{x}{2} \rfloor + m, \lfloor \frac{y}{2} \rfloor + n)},
		\label{eq5}
	\end{equation}
	where $(x, y)$ is the upper-left coordinate of the window and $(h, w)$ is the size of the window. As is known, entropy measures the uncertainty degree of information, namely positions with higher entropy scores are more likely to be real boundaries. Therefore, scoring through entropy can effectively localize boundary-around windows. 
	\begin{figure*}[!t]
		\centering
		\includegraphics[width=1\textwidth]{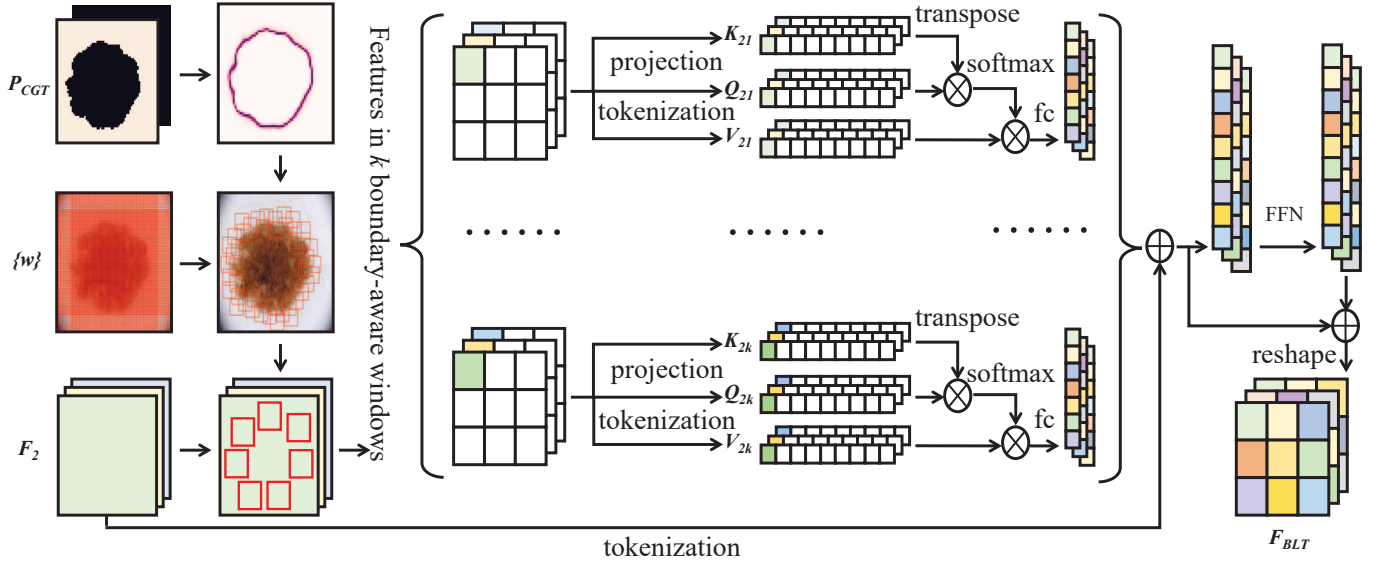}
		\caption{Overview of the boundary-aware local transformer (BLT) module. Long-range dependency is built within each window for fewer computational costs and less invalid information interference. Windows are generated adaptively around the object boundaries.}
		\label{fig4}
	\end{figure*}
	After performing non-maximum suppression \cite{r20} on $\{w\}$ according to the scores calculated by Eq. \ref{eq5}, a boundary-around window set $\{w^*\}$  and the corresponding feature set $\{f^*\}$ can be obtained by removing redundant windows and aligning $\{w^*\}$ with $F_2$ through RoIAlign \cite{r21}. Then, performing multi-head self-attention within each window's features $f^*_j \in f^*$ can build long-range dependencies by
	\begin{equation}	\begin{aligned}
			\mathcal{F}_{sa}(f^*_j) = softmax(\frac{f^*_jE_qf^*_jE_k}{\sqrt{d}})f^*_jE_v \\
			F_{sa} = (\mathcal{F}_{sa,1} \textcircled{c} \cdot \cdot \cdot \textcircled{c} \mathcal{F}_{sa,g}) \cdot W_{sa} \textcircled{+} F_2,
		\end{aligned}
		\label{eq6}
	\end{equation}
	where $E_q \in \mathbb{R} ^ {2C \times d}$, $E_k \in \mathbb{R} ^ {2C \times d}$, $E_v \in \mathbb{R} ^ {2C \times d}$, and $W_{sa} \in \mathbb{R} ^ {gd \times d}$ are learnable projection matrices. Finally, the output of BLT is produced by
	\begin{equation}
		F_{BLT} = (max(0, F_{sa}W_{s1}+b_{s1}) \cdot W_{s2}+b_{s2}) \textcircled{+} F_{sa},
		\label{eq9}
	\end{equation}
	where $W_{s1} \in \mathbb{R} ^ {d \times 4d} $ and  $W_{s2} \in \mathbb{R} ^ {4d \times d} $ are learnable projection matrices and $b_{s1}$, $b_{s2}$ $\in \mathbb{R}$ are the offsets in FFN. It should be noticed that only regions around boundaries will go through the above boundary-around local multi-head self-attention for representation learning. Thus, its computational complexity is
	\begin{equation}
		\Omega_{BLT} = k(6dhwC + 2d(hw)^2 + d^2hw),
		\label{eq7}
	\end{equation}
	where $k$ is the maximum number of windows. In our experiments, we adopt $h=\frac{H}{32}$, $w = \frac{W}{32}$, and set $k$ as $\alpha \frac{HW}{hw}, \alpha \in (0, 1) $. Then, Eq. \ref{eq7} is rewritten as
	\begin{equation}
		\begin{aligned}
			\Omega_{BLT}  &= \alpha \left( {6dHWC + \frac{2}{{32 \times 32}}d{{\left( {HW} \right)}^2} + {d^2}HW} \right) \\ &\approx {\rm O}\left( {\frac{\alpha }{{16 \times 32}}d{{\left( {HW} \right)}^2}} \right).
		\end{aligned}
		\label{eq7+}
	\end{equation}
	Compared to directly applying vanilla self-attention onto $F_2$ whose computational complexity is
	\begin{equation}
		\begin{aligned}
			\Omega_{ViT} &= \frac{3}{8}dHWC + \frac{1}{8}d(HW)^2 + \frac{1}{16}d^2HW \\ & \approx {\rm O}\left( {\frac{1}{{8}}d{{\left( {HW} \right)}^2}} \right),
			\label{eq8}
		\end{aligned}
	\end{equation}
	the computational complexity of BLT is at most $\times 64$ lower.
	
	\subsection{Multi-Scale Soft Supervision}
	
	For medical image segmentation using multi-scale features, inspired by \cite{r23}, we introduce soft supervision to alleviate boundary uncertainty brought by down-sampling or up-sampling.  Concretely, each ground-true mask $M \in \mathbb{R} ^ {H \times W} $ is resized to multiple scales for supervision (\textit{i.e.}, $G_{BLT}  \in \mathbb{R} ^ {c \times \frac{H}{2} \times \frac{W}{2}} $ for training BLT and $G_{CGT}  \in \mathbb{R} ^ {c \times \frac{H}{4} \times \frac{W}{4}} $ for training CGT) while preserving the original boundary distributions. Given class $l$ and the down-sampling scale $s$, the resized ground-true map $G_{l,s}$ is constructed as
	\begin{equation}
		G_{l,s}(i, j) = \frac{1}{{{2^{2(s - 1)}}}}\sum\limits_{(m,n) \in {O_{i,j}}} {\left| {M(m,n) =  = l} \right|},
		\label{eq10}
	\end{equation}
	where $O_{i,j}$ represents the down-sampling block in $M$ corresponding to $(i,j)$ in $G_{l,s}$ defined as $M((i-1)2^{s-1}+1:i2^{s-1}, (j-1)2^{s-1}+1:j2^{s-1})$. In this way, the probability value of $(i, j)$ in $G_{l,s}$ is determined by the frequency of class $l$ in the down-sampling block and the boundary distributions in $G_{l,s}$ would approach those in the original mask $M$. Given the down-sampling scale $s$, a set of resized ground-true maps are constructed accordingly as ${G_s} = \left\{ {{G_{l,s}}|l = 1,2,...,c} \right\}$. Then, the loss between ${P_s} = \left\{ {{P_{l,s}}|c = 1,2,...,c} \right\}$ and $G_s$ of the same scale is defined as
	\begin{equation}
		{\cal L}_s = \sum_{l=1}^{c}\sum_{i=1}^{\frac{H}{2^{s-1}}}\sum_{j=1}^{\frac{W}{2^{s-1}}}|G_{l,s}(i, j) - P_{l,s}(i, j)|.
		\label{eq11}
	\end{equation}
	
	After setting $s=1,2,3$ to define the loss functions ${\cal L}_{CNN}$, ${\cal L}_{BLT}$, and ${\cal L}_{CGT}$ for CNN, BLT, and CGT respectively, their features are further fused to generate the final predictions which are penalized by both a cross-entropy loss and a dice loss denoted as ${\cal L}_C$. The overall loss function ${\cal L}$ for multi-scale soft supervision of BATFormer becomes
	\begin{equation}
		{\cal L} = \beta_1 {\cal L}_{CNN} + \beta_2 {\cal L}_{BLT} + \beta_3 {\cal L}_{CGT} + \beta_4 {\cal L}_C,
		\label{eq12}
	\end{equation}
	where $\beta_1$, $\beta_2$, $\beta_3$, and $\beta_4$ are balancing hyper-parameters set as 0.2, 0.1, 0.1, and 0.6 respectively in our experiments.
	\begin{table*}[ht]
		\centering
		\caption{LV test results of BATFormer and the state-of-the-art CNN-based methods from the live ACDC leaderboard. Bold and underlined represent the best results of the non-ensemble and ensemble methods respectively.}\label{tab1}
		\begin{tabular}{ccccccccccc}
			\hline
			\multicolumn{1}{c}{\multirow{2}{*}{Method}} & \multicolumn{1}{c}{\multirow{2}{*}{Year}} & \multicolumn{2}{c}{Mean Dice}                   & \multicolumn{2}{c}{Mean Hausdorff}              & \multicolumn{2}{c}{Volume ES}                              & \multicolumn{2}{c}{Mass ED}      & \multirow{2}{*}{Ensemble}                          \\ \cline{3-10}
			\multicolumn{1}{c}{} & \multicolumn{1}{c}{}                      & \multicolumn{1}{c}{ED} & \multicolumn{1}{c}{ES} & \multicolumn{1}{c}{ED} & \multicolumn{1}{c}{ES} & \multicolumn{1}{c}{correlation} & \multicolumn{1}{c}{bias} & \multicolumn{1}{c}{correlation} & \multicolumn{1}{c}{bias}& \\ \hline
			Isensee \cite{r26}  &   2017                              & 0.967                  & 0.928                  & \underline{5.476}                & \underline{6.921}                  & 0.991                           & 0.490                    & 0.997                           & \underline{1.530}               & \Checkmark     \\
			Khened \cite{r30} & 2017                                  & 0.964                  & 0.917                  & 8.129                  & 8.968                  & 0.989                           & -0.548                   & \textbf{0.997}                           & \textbf{0.576}      & \XSolidBrush     \\
			Baumgartner \cite{r31}  &  2017                           & 0.963                  & 0.911                  & 6.526              & 9.170                  & 0.988                           & 0.568                    & 0.995                           & 1.436           	& \XSolidBrush         \\
			Wolterink \cite{r33}  &  2017                             & 0.961                  & 0.918                  & 7.515                & 9.603                  & 0.988                           & -0.494                   & 0.993                           & 3.046           	& \XSolidBrush         \\ 
			Zotti \cite{r27}  &  2018                                 & 0.964                  & 0.912                  & 6.180            & 8.386                  & 0.990                           & -0.476                   & \textbf{0.997}                           & 3.746           	& \XSolidBrush         \\
			Painchaud \cite{r28} &  2020                              & 0.961                  & 0.911                  & 6.152                  & 8.278                  & 0.990                           & -0.480                   & \textbf{0.997}                           & 3.824               & \XSolidBrush     \\
			Calisto \cite{r32}  &  2020                               & 0.958                  & 0.903                  & 5.592                & 8.644                  & 0.981                           & 0.494                    & 0.997                          & 3.072           	& \Checkmark         \\
			Simantiris \cite{r24} &  2020                             & 0.967                  & 0.928                  & 6.366                  & 7.573                  & \underline{0.993}                           & \underline{-0.360}                   & \underline{0.998}                  & 2.032               & \Checkmark     \\
			Girum \cite{r29} &  2021                                  & \textbf{0.968}         & 0.916                  & 6.422                 & 9.305                  & 0.985                           & 0.344                    & 0.997                           & -0.728          	& \XSolidBrush         \\
			Guo \cite{r25}  &  2022                                   & \underline{0.968}         & \underline{0.935}                  & 5.814                  & 7.361                  & 0.993                           & -0.552                   & \textbf{0.997}                           & 2.610               & \Checkmark     \\ \hline
			\textbf{BATFormer (ours)} &  2022                           & \textbf{0.968}         & \textbf{0.938}         & \textbf{5.456}         & \textbf{6.348}         & \textbf{0.994}                  & \textbf{0.210}            & \textbf{0.997}                           & \textbf{0.790}           & \XSolidBrush         \\
			\hline
		\end{tabular}
	\end{table*}
	\begin{table*}[ht]
		\centering
		\caption{RV test results of BATFormer and the state-of-the-art CNN-based methods from the live ACDC leaderboard. Bold and underlined represent the best results of the non-ensemble and ensemble methods respectively.}\label{tab2}
		\begin{tabular}{ccccccccccc}
			\hline
			\multicolumn{1}{c}{\multirow{2}{*}{Method}}& \multicolumn{1}{c}{\multirow{2}{*}{Year}} & \multicolumn{2}{c}{Mean Dice}                  & \multicolumn{2}{c}{Mean Hausdorff}              & \multicolumn{2}{c}{Volume ES}                              & \multicolumn{2}{c}{Mass ED}     & \multirow{2}{*}{Ensemble}                           \\ \cline{3-10}
			\multicolumn{1}{c}{}&\multicolumn{1}{c}{}                      & \multicolumn{1}{c}{ED} & \multicolumn{1}{c}{ES}  & \multicolumn{1}{c}{ED} & \multicolumn{1}{c}{ES} & \multicolumn{1}{c}{correlation} & \multicolumn{1}{c}{bias} & \multicolumn{1}{c}{correlation} & \multicolumn{1}{c}{bias}& \\ \hline
			Isensee \cite{r26}&2017                                   & 0.951                  & \underline{0.904}         & \underline{8.205}         & 11.655                 & \underline{0.910}                           & -3.750                   & 0.992                           & \underline{0.900}               & \Checkmark   \\
			Baumgartner \cite{r31} & 2017                              & 0.932                  & 0.883                  & 12.670                 & 14.691                 & 0.851                           & 1.218                    & 0.977                           & -2.290        & \XSolidBrush           \\
			Khened \cite{r30}      & 2017 & 0.935                  & 0.879                  & 13.994                 & 13.930                 & 0.858                           & -2.246                   & 0.982                           & -2.896        & \XSolidBrush           \\
			Zotti \cite{r43} & 2017      & 0.941                  & 0.882                  & 10.318                 & 14.053                 & 0.872                           & -2.228                   & 0.991                           & -3.722        & \XSolidBrush           \\
			Zotti \cite{r27} & 2018                                    & 0.934                  & 0.885                   & 11.052                 & \textbf{12.650}                 & 0.869                           & \textbf{-0.872}          & 0.986                           & 2.372         & \XSolidBrush           \\
			Simantiris \cite{r24} &2020                                & 0.936                  & 0.889                  & 13.289                 & 14.367                 & 0.894                           & \underline{-1.292}                   & 0.990                           & 0.906         & \Checkmark           \\
			Calisto \cite{r32} &2020                                  & 0.936                  & 0.884                  & 10.183                 & 12.234                 & 0.899                           & -2.118                   & 0.989                           & 3.550         & \Checkmark           \\
			Painchaud \cite{r28}   & 2020 & 0.933                  & 0.884                  & 13.718                 & 13.323                 & 0.865                           & -0.874                   & 0.986                           & 2.078         & \XSolidBrush           \\
			Girum \cite{r29} &2021                                    & 0.939                  & 0.893                  & 11.326                 & 13.306                 & \textbf{0.923}                  & -1.170                   & 0.978                           & 1.960         & \XSolidBrush           \\
			Guo \cite{r25}& 2022                                       & \underline{0.955}         & 0.894                  & 8.877                  & \underline{11.649}        & 0.879                           & -2.102                   & \underline{0.994}                           &   	2.112                  & \Checkmark         \\ \hline
			\textbf{BATFormer (ours)} & 2022                             & \textbf{0.948}                  & \textbf{0.903}                  & \textbf{9.272}                  & 13.357                 & 0.921                           & -2.230                   & \textbf{0.996}                  & \textbf{0.620}         & \XSolidBrush  \\
			\hline
		\end{tabular}
	\end{table*}
	\begin{table*}[ht]
		\centering
		\caption{Myo test results of BATFormer and the state-of-the-art CNN-based methods from the live ACDC leaderboard. Bold and underlined represent the best results of the non-ensemble and ensemble methods respectively.}\label{tab3}
		\begin{tabular}{ccccccccccc}
			\hline
			\multicolumn{1}{c}{\multirow{2}{*}{Method}}& \multicolumn{1}{c}{\multirow{2}{*}{Year}} & \multicolumn{2}{c}{Mean Dice}                   & \multicolumn{2}{c}{Mean Hausdorff}              & \multicolumn{2}{c}{Volume ES}                              & \multicolumn{2}{c}{Mass ED}      & \multirow{2}{*}{Ensemble}                          \\ \cline{3-10}
			\multicolumn{1}{c}{} & \multicolumn{1}{c}{}                       & \multicolumn{1}{c}{ED} & \multicolumn{1}{c}{ES} & \multicolumn{1}{c}{ED} & \multicolumn{1}{c}{ES} & \multicolumn{1}{c}{correlation} & \multicolumn{1}{c}{bias} & \multicolumn{1}{c}{correlation} & \multicolumn{1}{c}{bias}& \\ \hline
			Isensee \cite{r26} &2017                                 & 0.904                  & \textbf{0.923}          & \underline{7.014}                  	   & \underline{7.328}                  & 0.988                           & -1.984                   & 0.987                           & -2.547                 & \Checkmark     \\
			Baumgartner \cite{r31} & 2017                               & 0.892                  & 0.910                  & 8.703                     & 10.637                 & 0.983                           & -9.602                   & 0.982                           & -6.861                 & \XSolidBrush  \\ 
			Khened \cite{r30}      & 2017                              & 0.889                  & 0.898                  & 9.841                  		   & 12.582                 & 0.979                           & -2.572                   & \textbf{0.990}                           & -2.873                 & \XSolidBrush  \\
			Patravali \cite{r44}   & 2017                                   & 0.882                  & 0.897                  & 9.757                  	   & 11.256                 & 0.986                           & -4.464                   & 0.989                           & -11.586                & \XSolidBrush  \\
			Zotti \cite{r27} &2018                                    & 0.886                  & 0.902                  & 9.586                   & 9.291                  & 0.980                           & 1.160                    & 0.986                           & -1.827                 & \XSolidBrush  \\
			Painchaud \cite{r28} &2020                               & 0.881                  & 0.897                  & 8.651                  		   & 9.598                  & 0.979                           & 0.296                    & 0.987                           & -2.906                 & \XSolidBrush  \\
			Calisto \cite{r32} & 2020                                 & 0.873                  & 0.895                  & 8.197                  	   	   & 8.318                  & 0.988                           & \underline{-1.790}                   & 0.989                           & \underline{-2.100}                 & \Checkmark  \\
			Simantiris \cite{r24} & 2020                               & 0.891                  & 0.904                     & 8.264                  	   & 9.575                  & 0.983                           & -2.134                   & \underline{0.992}                  & -2.904                 & \Checkmark  \\
			Girum \cite{r29}       & 2021                              & 0.894                  & 0.906                  & 8.998                  	   & 9.922                  & 0.948                           & 0.920                    & 0.973                           & -0.754                 & \XSolidBrush  \\
			Guo \cite{r25} &2022                                   & \underline{0.906}         & \underline{0.923}            & 7.469                  	   & 7.702                  & \underline{0.989}                           & -2.880                   & 0.991                           & -2.388                 & \Checkmark     \\ \hline
			\textbf{BATFormer (ours)}&2022      & \textbf{0.895}                 			 & \textbf{0.916}                    & \textbf{6.944}         		  & \textbf{6.730}         & \textbf{0.991}         & \textbf{-0.25}           & 0.989                           & \textbf{-0.430}                 & \XSolidBrush\\
			\hline
		\end{tabular}
	\end{table*}
	
	\section{Evaluation}
	\label{sec:evaluation}
	
	In this section, we conduct extensive comparison experiments against the state-of-the-art CNN-based and transformer-based/-hybrid approaches on publicly-available datasets.
	
	\subsection{Datasets}
	
	The following two datasets covering 2D and 3D medical image data were adopted for evaluation:
	
	\subsubsection{ACDC}
	
	\begin{table*}[!t]
		\centering
		\caption{Comparison results with transformer-based methods on the ACDC dataset. \textit{P} represents the number of model parameters measured in millions. \textit{F} represents the floating-point operations per second measured in Giga, which is utilized to measure the computational complexity of models. Avg., RV, Myo, and LV represent the average, the right ventricle, the myocardium, and the left ventricle, respectively. The best results are marked in bold.}\label{tab4}
		\begin{tabular}{c|c|c|cccc|cccc|cc}
			\hline
			\multirow{2}{*}{Method Type} & \multirow{2}{*}{Method}& \multirow{2}{*}{Year} & \multicolumn{4}{c|}{Dice (\%) on All Samples} & \multicolumn{4}{c|}{Dice (\%) on Hard Samples} & \multirow{2}{*}{\textit{P}} & \multirow{2}{*}{\textit{F}}\\ \cline{4-11}
			&  &            &Avg. & RV & Myo & LV &Avg. & RV & Myo & LV & &\\ \hline
			\multirow{3}{*}{\makecell[c]{3D\\Transformer}} & nnFormer \cite{r40}&2021      & 92.06          & 90.94          & 89.58          & 95.65  &89.77 & 87.49 & 87.40 & 94.43    & 159 & 158    \\
			& UNETR \cite{r8}&2022        & 88.61          & 85.29          & 86.52          & 94.02  &86.38 & 81.53 & 84.75 & 92.84  & 92 & 86       \\
			& D-Former \cite{r41}&2022      & 92.29          & 91.33          & 89.60          & 95.93    & - &- & - & - & 44 & 54      \\ \hline
			\multirow{6}{*}{\makecell[c]{2D\\Transformer}} & SETR-PUP$^*$ \cite{r5_4}&2021 & 85.56          & 82.52          & 81.57          & 92.60  & 85.56 & 82.52 &81.57 & 92.60 & 40 & 39     \\
			& TransUNet$^\diamondsuit$ \cite{r6}&2021     & 89.71          & 88.86          & 84.54          & 95.73  & 89.05 & 87.84 & 84.93 & 94.38  & 105 & 25       \\
			& SwinUnet$^\diamondsuit$ \cite{r10}&2021    & 90.00          & 88.55          & 85.62          & 95.83 & 89.35 & 87.82 & 85.58 & 94.65 & 41  & 12       \\
			& MISSFormer$^\diamondsuit$ \cite{r15}&2021     & 90.86          & 89.55          & 88.04   & 94.99     & 89.81 & 87.56 & 86.98 & 94.90  & 42 & 7.2       \\
			& MedT$^*$ \cite{r7}&2021      & 90.98          & 89.18          & 88.39          & 95.37  & 89.30 & 86.38 & 86.81 & 94.70     & 1.6 & 4.4   \\
			& \textbf{BATFormer (ours)}&2022     & \textbf{92.84} & \textbf{91.97} & \textbf{90.26} & \textbf{96.30}  & \textbf{91.14} & \textbf{87.99} & \textbf{89.48} & \textbf{95.97} & \textbf{1.2} & \textbf{4.1} \\ \hline
		\end{tabular}
		\begin{tablenotes}
			\centering
			\item $*$ represents the re-implemented results based on the released code. 
			\item $^\diamondsuit$ represents the results on all samples from the released paper and results on hard samples from the released code.
		\end{tablenotes}
	\end{table*}
	The ACDC dataset \cite{r19} consists of short-axis cine-MRI from 150 patients acquired at the University Hospital of Dijon. Structures of interest, including left ventricle (LV), right ventricle (RV), and myocardium (Myo) were annotated manually by experienced experts on end-diastolic (ED) and end-systolic (ES) phase instants. In the official challenge, the ACDC dataset was divided into 100 and 50 patients for training and testing respectively. The ground-true masks of the training set are accessible, while the predictions of the testing set must be submitted to the online evaluation platform for performance evaluation.
	
	As the state-of-the-art CNN-based and transformer-based/-hybrid approaches adopted different data partitioning methods for evaluation, we compared BATFormer with the CNN-based methods by following the official data split and submitting the results to the online evaluation platform and with the transformer-based/-hybrid approaches by following the setting in \cite{r6} (\textit{i.e.}, the first transformer-based work for medical image segmentation where the 100 patients were divided into 7:1:2 for training, validation, and testing respectively.
	
	\subsubsection{ISIC 2018}
	
	The ISIC dataset \cite{r17, r18} contains a total of 2596 dermoscopic images with well-annotated labels. As the state-of-the-art methods on this dataset, including both CNN-based \cite{r35, r36} and transformer-based \cite{r38}, adopted different data splits, for a fair comparison, we randomly divided the ISIC 2018 dataset into 2076 images for training and 520 images for testing and re-implemented the comparison methods according to their open source codes.
	
	\subsection{Implementation Details}
	
	All the networks were trained using an Adam optimizer with an initial learning rate of 0.001 and a batch size of 4. All methods were implemented within the PyTorch framework and trained on NVIDIA 3090ti GPUs for 400 rounds. To better fuse multi-scale features, in the last 50 rounds, the parameters of the backbone were frozen, and only the parameters of the branch heads and the synthetic head were updated with a learning rate of $10^{-4}$. Random rotation, random scaling, cropping, contrast adjustment, and gamma augmentation were applied for data augmentation.
	For the ACDC dataset, following the same practice of \cite{r6}, \cite{r10}, and \cite{r15}, all patient-level 3D images are preprocessed into slice-level 2D images for training and the metrics are calculated at the patient level for a fair comparison.
	\begin{figure*}[!t]
		\centering
		\includegraphics[width=1\textwidth]{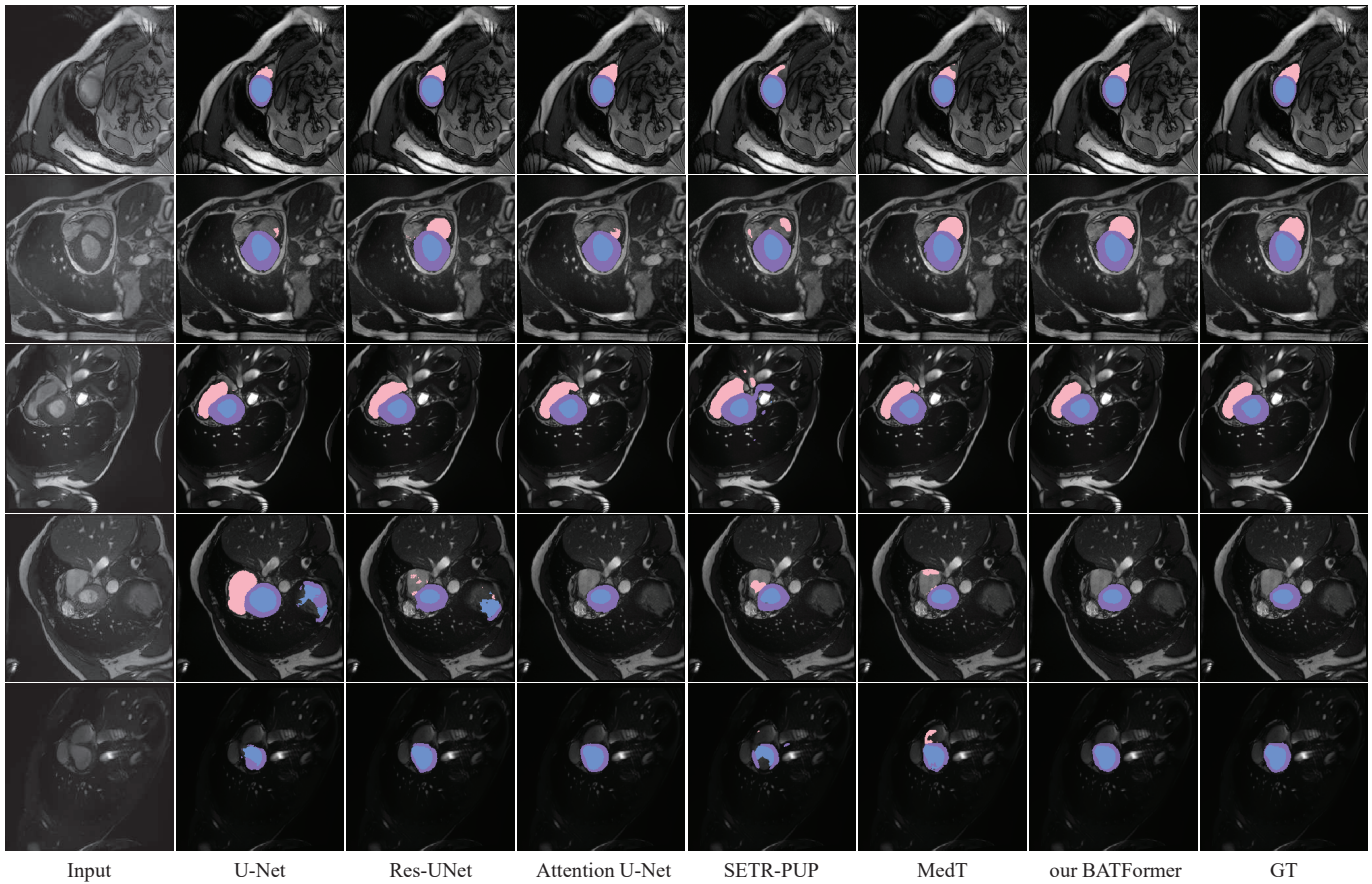}
		\caption{Qualitative comparison results on the ACDC dataset. From left to right: the raw images, the segmentation maps produced by U-Net \cite{r1}, Res-UNet \cite{r2}, Attention U-Net \cite{r3}, SETR-PUP \cite{r5_4}, MedT \cite{r7}, and our BATFormer respectively, and the manual annotations. The pink, purple, and blue regions denote the right ventricle, the myocardium, and the left ventricle respectively.}
		\label{fig7}
	\end{figure*}
	\begin{figure*}[!t]
		\centering
		\includegraphics[width=1\textwidth]{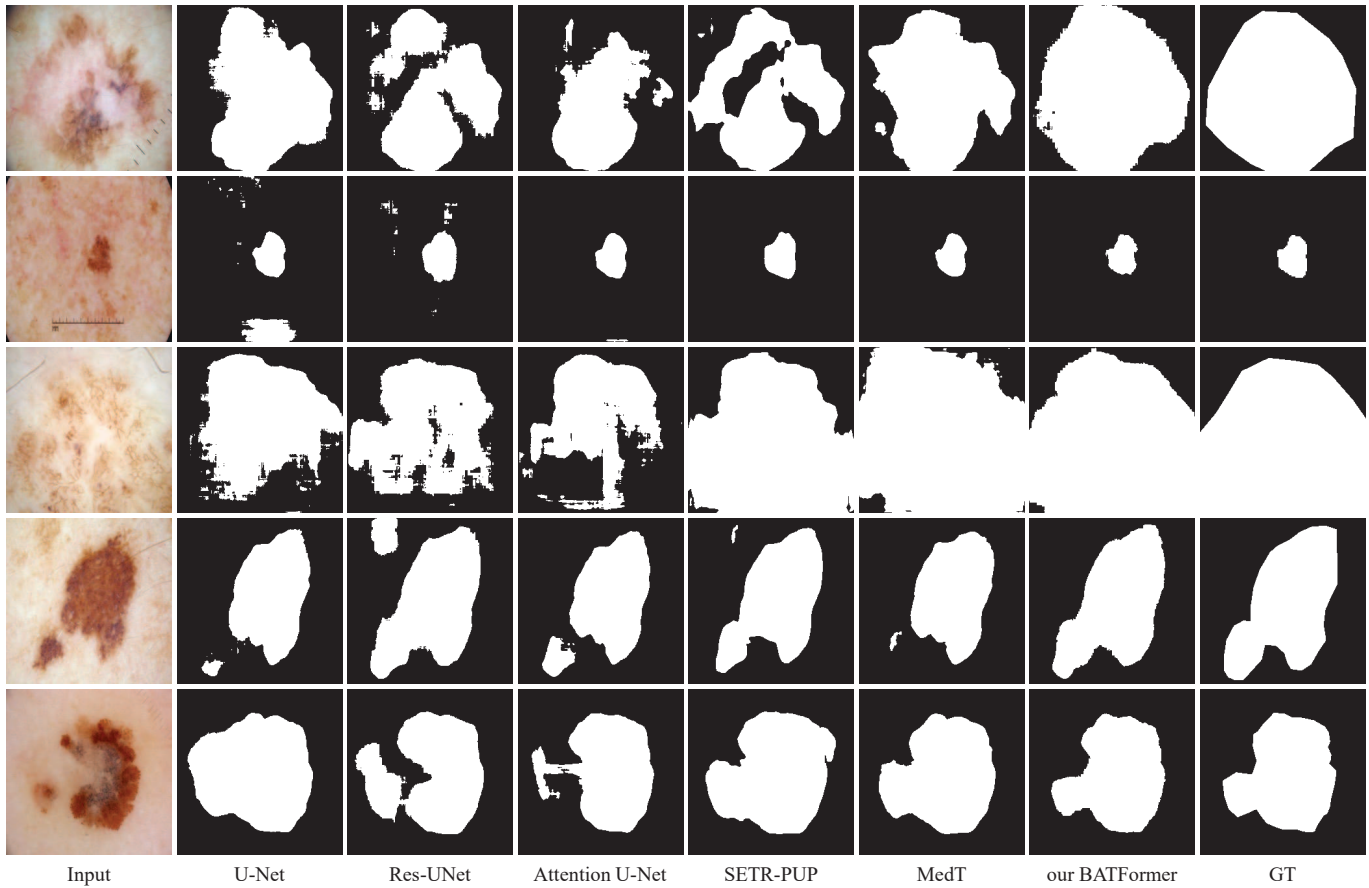}
		\caption{Qualitative comparison results on the ISIC 2018 dataset. From left to right: the raw images, the segmentation maps produced by U-Net \cite{r1}, Res-UNet \cite{r2}, Attention U-Net \cite{r3}, SETR-PUP \cite{r5_4}, MedT \cite{r7}, and our BATFormer respectively, and the manual annotations.}
		\label{fig5}
	\end{figure*}
	\begin{table*}[!t]
		\caption{Comparison results on the ISIC 2018 dataset. \textit{P} represents the number of model parameters measured in millions. \textit{F} represents the floating-point operations per second measured in GFLOP/s, which is utilized to measure the computational complexity of models. The best results are marked in bold.}\label{tab5}
		\centering
		\begin{tabular}{c|c|c|ccccc|ccccc|cc}
			\hline
			\multirow{2}{*}{Method Type} & \multirow{2}{*}{Method}& \multirow{2}{*}{Year} & \multicolumn{5}{c|}{Evaluation on All Samples (\%)}   & \multicolumn{5}{c|}{Evaluation on Hard Samples (\%)}  & \multirow{2}{*}{\textit{P}} & \multirow{2}{*}{\textit{F}} \\ \cline{4-13}
			&       &                  & Dice           & IoU            & ACC            & SE             & SP    & Dice           & IoU            & ACC            & SE             & SP         & \multicolumn{1}{l}{}     & \multicolumn{1}{l}{}                        \\ \hline		\multirow{6}{*}{CNN}         & U-Net \cite{r1} & 2015                  & 88.18          & 80.89          & 95.50          & 89.41          & 97.44  & 76.84          & 64.89          & 91.33          & 78.53          & 96.06        & 31                                              & 48                                          \\
			& Attention U-Net \cite{r3} &2019              & 88.30          & 80.84          & 95.28          & 89.62          & 97.46   & 77.39          & 65.01          & 90.93       & 80.10       & 96.36       & 35                                              & 67                                          \\
			& CE-Net \cite{r39}&2019                   & 89.58          & 82.71          & 96.00          & 90.45          & 97.75      & 78.56          & 66.68          & 92.64          & 80.18        & 96.52    &   29                                              & 7.2                                         \\
			& CA-Net \cite{r37} &2020                & 89.46          & 82.61          & 96.04          & 91.21          & 96.70      & 78.53          & 66.74          & 92.88          & 81.21          & 95.39    &  2.8                                             & 5.6                                         \\
			& CPFNet \cite{r36} &2020               & 89.88          & 82.97          & 96.00          & 91.02          & 97.15     & 79.51          & 67.69          & 92.05          & 81.55          & 95.66     & 43                                           & 8.1                                         \\
			& nnU-Net \cite{r42} &2021              & 90.16          & 83.79          & 96.40          & 91.02          & 97.55    & 78.16          & 65.96          & 92.69          & 80.76          & 96.60       & 31                                              & 48                                         \\
			& Ms RED \cite{r35} &2022             & 90.25          & 83.59          & 96.46          & \textbf{91.94} & 97.36      & 79.48          & 67.62          & 93.19       & 83.43 & 96.03     &   4.7                                             & 9.0                                         \\ \hline
			\multirow{5}{*}{Transformer} & SETR-PUP \cite{r5_4}&2021                & 88.03          & 80.53          & 95.51          & 91.51          & 96.52  & 77.39          & 65.46          & 91.91          & 84.28          & 94.68         & 40                                              & 39                                          \\
			& TransUNet \cite{r6}&2021              & 88.88          & 81.85          & 95.94          & 90.08          & 97.89       & 76.49          & 64.17          & 92.21         & 79.83          & \textbf{96.99}    & 105                                             & 25                                          \\
			& MedT \cite{r7}&2021        & 89.28          & 82.48          & 95.83          & 88.16          & \textbf{98.00}   & 77.36          & 65.38          & 91.55          & 75.79         & 96.67 &  1.6                                             & 4.4                                         \\
			& FAT-Net \cite{r38}&2022              & 89.08          & 82.08          & 95.85          & 90.46          & 97.34    & 77.14          & 65.19          & 91.82          & 80.03          & 96.37      & 30                                              & 23                                          \\
			& \textbf{BATFormer (ours)}&2022       & \textbf{90.76} & \textbf{84.64} & \textbf{96.76} & 91.22          & 97.74     & \textbf{81.81} & \textbf{70.45} & \textbf{95.28} & \textbf{85.10}      & 95.87     & \textbf{1.2}                                    & \textbf{4.1}                                \\  \hline 
		\end{tabular}
	\end{table*}
	
	\subsection{Evaluation on ACDC}
	
	\subsubsection{Learning Frameworks for Comparison}
	For evaluation on the 3D ACDC dataset, the top six CNN-based methods on the live ACDC test leaderboard,  2D transformer-based methods including SETR \cite{r5_4}, MISSFormer \cite{r15}, TransUNet \cite{r6}, SwinUnet \cite{r10}, and MedT\cite{r7}, and 3D transformer-based methods including UNETR \cite{r8}, nnFormer \cite{r40}, and D-Former \cite{r41} are used for comparison.
	
	\subsubsection{Quantitative Results}
	
	Quantitative results from the online live ACDC leaderboard are summarized in Tables \ref{tab1}, \ref{tab2}, and \ref{tab3}. Here, methods ranking the top ten in each category of the ACDC live leaderboard are included for comparison. It should be noted that all the state-of-the-art approaches are CNN-based. For each category, BATFormer ranks top three of all approaches, achieving the best overall segmentation performance among the non-ensemble models. In addition, BATFormer's performance is quite close to that of non-ensemble approaches but is much more lightweight and applicable in clinical scenarios.
	
	Quantitative results of transformer-based methods on the ACDC dataset are summarized in Table~\ref{tab4}. In general, 3D transformer-based methods can achieve better segmentation performance with the help of inter-slice spatial information. Specifically, D-Former has the best performance with an average of $92.29\%$ in Dice.
	Compared to the state-of-the-art 3D transformers, BATFormer achieves considerable performance improvements on the ACDC dataset, outperforming D-Former by an average of $0.54\%$ in Dice with much lower computational complexity. In addition, BATFormer outperforms other 2D transformer-based approaches by a large margin, demonstrating its effectiveness in capturing rich global features with less computational complexity by CGT and accurate boundary detection by BLT.
	
	\subsubsection{Qualitative Results}
	
	Qualitative segmentation results of different methods on the ACDC dataset are presented in Fig.~\ref{fig7}. Suffering from local contextual similarity, CNN-based methods produce more false positives. Comparatively, transformer-based methods produce extensive false negatives with distorted boundaries due to rigid partitioning. Thanks to both CGT and BLT, BATFormer achieves the best segmentation performance, especially on boundary consistency with manual annotations.
	
	\subsection{Evaluation on ISIC 2018}
	
	\subsubsection{Learning Frameworks for Comparison}
	
	Both state-of-the-art CNN-based and transformer-based/-hybrid architectures have been included for comparison on the ISIC 2018 dataset. CNN-based architectures include U-Net \cite{r1}, Attention U-Net \cite{r3}, 
	CE-Net \cite{r39}, CA-Net \cite{r37}, 
	CPFNet\cite{r36}, nnU-Net \cite{r42}, and Ms RED \cite{r35}, while transformer-based approaches include SETR \cite{r5_4}, TransUNet \cite{r6}, MedT \cite{r7}, and FAT-Net \cite{r38}. Among these methods, Ms RED and FAT-Net are specifically designed for skin legion segmentation, SETR is for semantic segmentation, and the rest approaches are the most recent/representative medical image segmentation works. 
	
	\subsubsection{Quantitative Results}
	
	Quantitative comparison results of different methods on the ISIC 2018 dataset are summarized in Table~\ref{tab5}. Among CNN-based methods, Ms RED achieves the best performance while MedT achieves the best performance among transformer-based methods. Compared to the CNN-based methods, well-designed transformer-based methods achieve competitive segmentation performance with fewer parameters and lower FLOPs. Compared to Ms RED, BATFormer achieves consistent performance improvements by an average increase of 0.51$\%$, 1.05$\%$, 0.3$\%$, and 0.38$\%$ in Dice, IoU, ACC, and SP respectively. As for MedT, BATFormer outperforms it by an average increase of 1.48$\%$, 2.16$\%$, 0.93$\%$, and 3.06$\%$ in Dice, IoU, ACC, and SE respectively. Though Ms RED and MedT own the best SE and SP results, their overall performance is worse than that of BATFormer. Furthermore, BATFormer is the most lightweight framework with the fewest parameters and the lowest FLOPs, indicating its easy-to-deployment property.
	
	\subsubsection{Qualitative Results}
	
	Exemplar qualitative results produced by different approaches on the ISIC 2018 dataset are provided in Fig.~\ref{fig5}. Given local contextual similarity as shown in the second row of Fig.~\ref{fig5}, CNN-based methods can hardly distinguish skin lesions from the background resulting in extensive false positives, while transformer-based methods can recognize skin lesions correctly. In terms of large-scale skin lesions, as shown in the third row of Fig.~\ref{fig5}, segmentation maps generated by transformer-based methods are more complete than those produced by the CNN-based methods, validating the inadequate receptive fields of CNN. Compared to the state-of-the-art transformer-based methods, BATFormer effectively reduces false negatives with better boundary preservation, benefiting from both CGT and BLT respectively.
	
	\subsection{Evaluation on Hard Samples}
	
	For a more comprehensive evaluation, we further validate the effectiveness of BATFormer on hard samples which are defined as those samples whose Dice scores are lower than the average Dice score of the second-best method (\textit{i.e.}, 92.29\% on the ACDC dataset and 90.25\% on the ISIC 2018 dataset) according to \cite{r45}. Quantitative comparison results on the ISIC 2018 dataset are summarized in Table~\ref{tab5}. Among all comparison methods, CPFNet achieves the best overall performance with Dice, IoU, ACC, SE, and SP of $79.51\%$, $67.69\%$, $92.05\%$, $81.55\%$, and $95.66\%$ respectively. Compared to CPFNet, BATFormer achieves consistent performance improvements with an average improvement of 2.3$\%$, 2.76$\%$, 3.23$\%$, 3.55$\%$, and 0.21$\%$ in Dice, IoU, ACC, SE, and SP respectively. Though BATFormer achieves slightly lower SP, its Dice, IoU, ACC, and SE scores significantly outperform all comparison methods. In terms of the ACDC dataset, as summarized in Table~\ref{tab4}, BATFormer stably outperforms other comparison methods across all evaluation metrics. The prominent performance of BATFormer on hard samples further demonstrates its effectiveness.
	
	\section{Discussion}
	\label{sec:discussion}
	
	\subsection{Effectiveness of Each Component in BATFormer}
	
	\begin{figure}[!t]
		\centering
		\includegraphics[width=1\columnwidth]{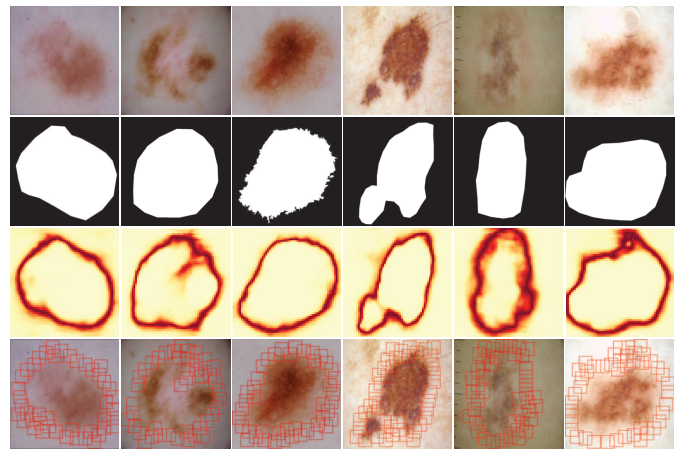}
		\caption{Visualization of boundary-around windows generation by BLT on the ISIC 2018 dataset. From top to bottom: the raw images, the ground truth, the entropy maps, and the generated boundary-aware windows. In entropy maps, the darker the color, the higher the entropy.}
		\label{fig8}
	\end{figure}
	\begin{table}[!t]
		\centering
		\caption{Ablation study on different component combinations of BATFormer on the ISIC 2018 dataset. BATFormer- represents the BATFormer without multi-scale soft supervision. The best results are marked in bold.}\label{tab6}
		\begin{tabular}{c|ccccc}
			\hline
			\multirow{2}{*}{Method} & \multicolumn{5}{c}{Evaluation Metrics (\%)}                                        \\ \cline{2-6}
			& Dice           & IoU            & ACC           & SE             & SP             \\ \hline
			backbone                & 88.41          & 81.09          & 95.53          & 89.46          & 97.51          \\
			+ CGT                   & 89.60          & 82.84          & 96.07          & 90.85          & 97.48          \\
			+ BLT                   & 89.73          & 83.30          & 96.36          & \textbf{91.55} & 97.47          \\
			BATFormer-              & 90.00          & 83.45          & 96.18          & 90.78          & 96.68           \\
			BATFormer                & \textbf{90.76} & \textbf{84.64} & \textbf{96.76} & 91.22          & \textbf{97.74} \\ \hline
		\end{tabular}
	\end{table}
	We evaluate the effectiveness of CGT and BLT by separately introducing them to the backbone network (\textit{i.e.} multi-scale U-Net) for medical image segmentation on the ISIC 2018 dataset. Quantitative comparison results summarized in Table~\ref{tab6} indicate that coupling either component of BATFormer is helpful for performance improvement, which is consistent with the analysis in Section II. Specifically, by capturing global dependency based on cross-scale features, CGT improves the overall segmentation performance by an average increase of 1.19 $\%$, 1.75$\%$, 0.54$\%$, and 1.21$\%$ in Dice, IoU, ACC, and SE respectively.  Comparatively, BLT achieves greater performance improvements, achieving an average increase of 1.32 $\%$, 2.21$\%$, 0.83$\%$, and 2.09$\%$ in Dice, IoU, ACC, and SE respectively. This can be explained by the fact that boundary uncertainty is more common and challenging in medical image segmentation, addressing which would be relatively more beneficial. In addition to CGT and BLT, we further compare the segmentation performance of BATFormer with and without multi-scale soft supervision as summarized in Table~\ref{tab6}. In general, introducing multi-scale soft supervision achieves consistent performance improvements across all metrics.
	\begin{figure}[!t]
		\centering
		\includegraphics[width=1\columnwidth]{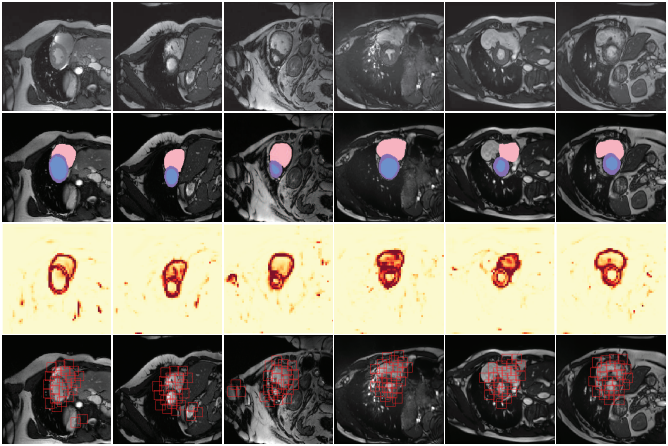}
		\caption{Visualization of boundary-around windows generation by BLT on the ACDC datasets. From top to bottom: the raw images, the ground truth, the entropy maps, and the generated boundary-aware windows. In entropy maps, the darker the color, the higher the entropy.}
		\label{fig8_2}
	\end{figure}
	\begin{table}[!t]
		\centering
		\caption{Ablation study on different window sizes of the BLT module on the ISIC 2018 dataset. The best results are marked in bold.}\label{tab7}
		\begin{tabular}{c|ccccc}
			\hline
			\multirow{2}{*}{Window size} & \multicolumn{5}{c}{Evaluation Metrics (\%)}                       \\ \cline{2-6}
			& Dice        & IoU         & ACC        & SE         & SP         \\ \hline
			8x8         & 89.66          & 83.05          & 96.24          & 90.35          & \textbf{97.87} \\
			12x12       & 89.59          & 82.88          & 96.32          & \textbf{92.07} & 97.11          \\
			16x16       & \textbf{89.73} & \textbf{83.30} & \textbf{96.36} & 91.55          & 97.47          \\
			20x20       & 89.55          & 82.97          & 96.30          & 90.99          & 97.63          \\ \hline
		\end{tabular}
	\end{table}
	
	To better evaluate the effectiveness of BLT, we further plot the generated entropy maps and boundary-around windows as shown in Figs. \ref{fig8} and \ref{fig8_2}. In general, boundaries are effectively localized through entropy calculation, based on which BLT can build boundary-specific mid-range dependency for performance improvement. 
	
	\subsection{Effect of Window Size in BLT}
	
	To analyze the effect of the window size $(h, w)$ in BLT, we evaluate the performance of the baseline multi-scale U-Net coupled with BLT under different window sizes $(h, w) = (8, 8)$, $(12, 12)$, $(16, 16)$, and $(20, 20)$. Quantitative comparison results are provided in Table~\ref{tab7}. With smaller window sizes, more windows would be regarded as boundary windows, resulting in redundancy and poor mid-range dependency establishment. Comparatively, using larger window sizes can better capture mid-range dependency within each window for boundary identification but may fail to preserve complete boundaries. Therefore, the selection of $(h, w)$ is task-specific. According to the comparison results in Table~\ref{tab7}, setting $(h, w)$ as $(16, 16)$ achieves the best BLT performance on the ISIC 2018 dataset. It should be noted that the performance gaps under various $(h, w)$ settings are quite limited, demonstrating the robustness of BLT in boundary preservation.
	\begin{table}[!t]
		\centering
		\caption{Ablation study on feature map selection of CGT on the ISIC 2018 dataset. Backbone-d5 represents the backbone with five downsampling blocks. The best results are marked in bold.}\label{tab8}
		\begin{tabular}{c|ccccc}
			\hline
			\multirow{2}{*}{Method} & \multicolumn{5}{c}{Evaluation Metrics (\%)}                                        \\ \cline{2-6}
			& Dice           & IoU            & ACC           & SE             & SP             \\ \hline
			backbone-d5                & 88.65          & 81.48          & 95.73          & 91.15          & 97.02          \\
			+ CGT-$F_3F_4F_5$                   & 89.55          & 82.72          & 96.03          & \textbf{91.42}          & 97.07          \\
			+ CGT-$F_3F_5F_6$                   & 89.50          & 82.59          & 96.06          & 90.07 & \textbf{97.51}          \\
			+ CGT-$F_3F_4F_5F_6$                & \textbf{89.87} & \textbf{83.12} & \textbf{96.11} & 91.27          & 97.04 \\ \hline
		\end{tabular}
	\end{table}
	
	\subsection{Effect of Feature Map Selection in CGT}
	
	To validate the effectiveness of CGT with various feature maps, we conducted ablation studies on a deeper backbone with five downsampling blocks coupled with CGT under different combinations ($F_3$, $F_4$, $F_5$), ($F_3$, $F_5$, $F_6$), and ($F_3$, $F_4$, $F_5$, $F_6$) respectively. In all experiments, $F_3$ is utilized to generate multi-group queries while $F_4$, $F_5$, and $F_6$ are utilized to generate the corresponding keys and values. Then, multi-group query-key-value outputs are combined and fed into the following feed-forward networks. Quantitative comparison results provided in Table~\ref{tab8} indicate that CGT achieves stable performance improvements while being insensitive to feature map selection. Specifically, CGT coupled with all deep semantic features (\textit{i.e.}, $F_3$, $F_4$, $F_5$, and $F_6$) achieves the best results, indicating that introducing richer semantic features is more beneficial for CGT.
	\begin{table}[!t]
		\centering
		\caption{Ablation study on feature map selection of BLT on the ISIC 2018 dataset. The best results are marked in bold.}\label{tab9}
		\begin{tabular}{c|ccccc}
			\hline
			\multirow{2}{*}{Method} & \multicolumn{5}{c}{Evaluation Metrics (\%)}                                        \\ \cline{2-6}
			& Dice           & IoU            & ACC           & SE             & SP             \\ \hline
			backbone                & 88.41          & 81.09          & 95.53          & 89.46          & 97.51          \\
			+ BLT-$F_1$                   & 89.57          & 82.67          & 96.20          & 90.57          & 97.33          \\
			+ BLT-$F_2$                   & \textbf{89.73}          & \textbf{83.30}          & \textbf{96.36}          & \textbf{91.55} & 97.47          \\
			+ BLT-$F_3$                   & 89.53          & 82.73          & 96.21          & 89.76 & \textbf{97.84}          \\
			+ BLT-$F_4$                   & 89.41          & 82.32          & 96.04          & 91.21 & 97.24          \\
			+ BLT-$F_5$                & 89.13          & 82.13          & 96.02          & 90.96  & 97.35 \\ \hline
		\end{tabular}
	\end{table}
	
	\subsection{Effect of Feature Map Selection in BLT}
	
	We analyze the effect of feature map selection in BLT by deploying it onto various feature maps. As summarized in Table~\ref{tab9}, coupling BLT with any feature map improves the performance. Among all feature maps, $F_2$ is the most suitable feature map for BLT, leading to the highest Dice of $89.73\%$. Comparatively, applying BLT onto $F_5$ is the least beneficial, $0.57\%$ lower in Dice than that of $F_2$. The above results are consistent with the observation that local fine features in shallow feature maps are more helpful for accurate boundary localization. It should be noted that coupling BLT with $F_1$ is not as beneficial as with $F_2$. One possible reason is that building long-range dependency on $F_1$ may result in redundancy which in turn degrades the performance of BLT.
	\begin{table}[!t]
		\centering
		\caption{Comparison results of CGT based on different kinds of transformers. CG-BLT represents the combination of BLT and CGT. The best results are marked in bold.}\label{tab10}
		\begin{tabular}{c|ccccc}
			\hline
			\multirow{2}{*}{Method} & \multicolumn{5}{c}{Evaluation Metrics (\%)}                                        \\ \cline{2-6}
			& Dice           & IoU            & ACC           & SE             & SP             \\ \hline
			backbone                & 88.41          & 81.09          & 95.53          & 89.46          & \textbf{97.51}          \\
			+ CGT                 & \textbf{89.60}          & \textbf{82.84}          & \textbf{96.07}          & \textbf{90.85}          & 97.48          \\
			+ CG-BLT                & 89.22 & 82.22 & 95.90 & 90.43          & 97.30 \\ \hline
		\end{tabular}
	\end{table}
	
	\subsection{Embedding BLT into CGT}
	
	To explore the potential combination of BLT and CGT, we introduce BLT to CGT and deploy them on $F_3$ of the backbone. Quantitative results are summarized in Table~\ref{tab10}. Though BLT can be successfully included in CGT, which outperforms the backbone, the performance gain is lower than that of the vanilla CGT as important global dependency may be blocked by BLT.  
	\begin{table}[!t]
		\centering
		\caption{Ablation study on hyper-parameters \{$\beta$\} on the ISIC 2018 dataset. The best results are marked in bold.}\label{tab11}
		\begin{tabular}{c|ccccc}
			\hline
			\multirow{2}{*}{($\beta_1,\beta_2,\beta_3,\beta_4$)} & \multicolumn{5}{c}{Evaluation Metrics (\%)}                        \\ \cline{2-6}
			& Dice           & IoU            & ACC           & SE             & SP             \\ \hline
			(0.10,0.10,0.10,0.70)                & 90.56          & 84.15          & 96.65          & 90.48          & 97.60          \\
			(0.20,0.10,0.10,0.60)                 & \textbf{90.76} & \textbf{84.64} & \textbf{96.76} & \textbf{91.22}          & \textbf{97.74}          \\
			(0.20,0.20,0.20,0.40)                 & 90.41          & 83.87          & 96.56          & 91.14          & 97.23          \\
			(0.25,0.25,0.25,0.25)                & 90.07          & 83.63          & 95.99          & 90.53         & 96.19 \\ \hline
		\end{tabular}
	\end{table}
	
	\subsection{Effect of Hyper-parameters \{$\beta$\}}
	
	We explore the effect of hyper-parameters \{$\beta$\} = ($\beta_1,\beta_2,\beta_3,\beta_4$) by training BATFormer under different settings, \textit{i.e.}, ($\beta_1,\beta_2,\beta_3,\beta_4$) = (0.1,0.1,0.1,0.7), (0.2,0.1,0.1,0.6), (0.2,0.2,0.2,0.4), and (0.25,0.25,0.25,0.25). Quantitative comparison results are summarized in Table~\ref{tab11}. Decreasing the sum of $\beta_1$ and $\beta_4$ would significantly degrade the overall performance. In general, BATFormer is insensitive to \{$\beta$\} except when the weight of $\beta_4$ is relatively low. In practice, for optimal training, we suggest setting ($\beta_1,\beta_2,\beta_3,\beta_4$) as (0.1,0.1,0.1,0.7) first, and then decreasing $\beta_4$ progressively while increasing $\beta_1$, $\beta_2$, and $\beta_3$ orderly. It should be noted that the sum of \{$\beta$\} should be kept as one during tuning.
	\begin{table}[!t]
		\centering
		\caption{Comparison results of BATFormer under 2D and 3D settings on the ACDC dataset. \textit{P} represents the number of model parameters measured in millions. \textit{F} represents the floating-point operations per second measured in GFLOPS/s, which is utilized to measure the computational complexity of models. The best results are marked in bold.}\label{tab12}
		\begin{tabular}{c|cccc|cc}
			\hline
			\multirow{2}{*}{Method Type} & \multicolumn{4}{c|}{Dice (\%)} & \multirow{2}{*}{\textit{P}} & \multirow{2}{*}{\textit{F}}\\  \cline{2-5}
			&Avg. & RV & Myo & LV & &\\ \hline 
			2D BATFormer & \textbf{92.84} & 91.97 & \textbf{90.26} & \textbf{96.30}  & \textbf{1.2} & \textbf{4.1}       \\
			3D BATFormer & \textbf{92.84} & \textbf{93.06} & 89.46 & 96.00  & 6.2 & 73  \\ \hline
		\end{tabular}
	\end{table}
	
	\subsection{Extension to 3D}
	
	BATFormer is extendable to 3D by simply replacing 2D convolutions with 3D in the backbone and tokenizing 3D feature maps into voxel sequences in transformers. According to the comparison results summarized in Table~\ref{tab12}, 3D BATFormer achieves comparable performance as 2D BATFormer on the ACDC dataset while demanding more parameters and computational resources. This conclusion is consistent with the quantitative results discussed in Section IV.C.
	
	\subsection{Model Efficiency Analysis}
	
	\begin{figure*}[!t]
		\centering
		\includegraphics[width=1\textwidth]{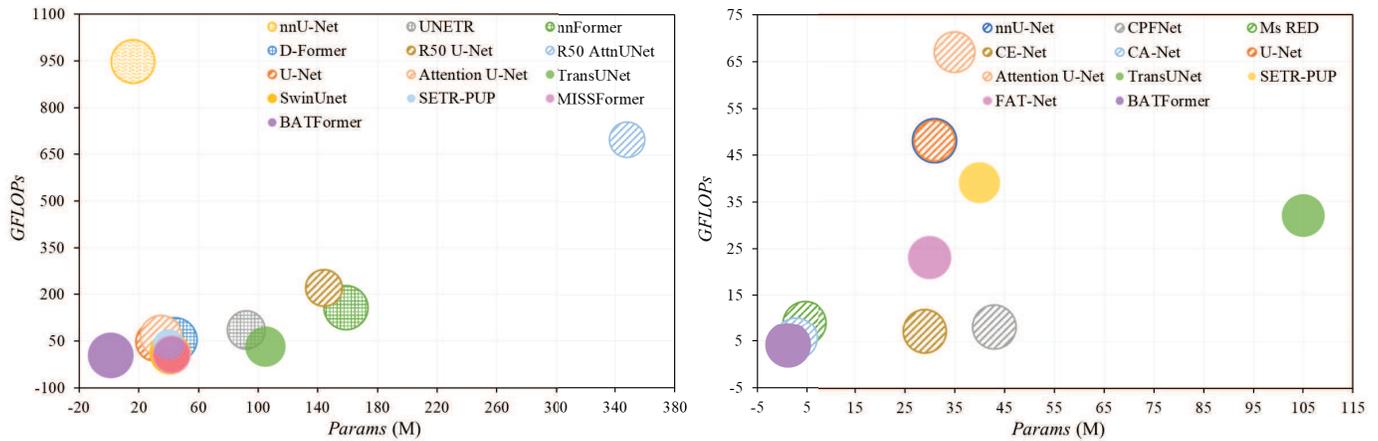}
		\caption{Model parameters and computational complexity comparison of different methods on the ACDC (\textbf{left}) and the ISIC 2018 (\textbf{right}) datasets. Circles filled with grid, horizontal lines, slashes, and solid colors indicate the 3D transformer-related, the 3D CNN-based, the 2D CNN-based, and the 2D transformer-related methods respectively. The larger the circle is, the better performance the corresponding model achieves. \textit{Params} represents the number of model parameters measured in millions (M). \textit{FLOPs} is the abbreviation of floating-point operations per second, which is utilized to measure the computational complexity of models, and \textit{GFLOPs} is short for \textit{Giga FLOPs}.}
		\label{fig9}
	\end{figure*}
	To evaluate the deployment feasibility of both CNN-based and transformer-based/-hybrid approaches, we compare their model efficiency as illustrated in Fig. \ref{fig9}. For each circle, the closer it approaches the bottom-left corner, the more lightweight the corresponding method is for deployment in realistic applications.
	On the 3D ACDC dataset, nnU-Net requires the most computations and R50 AttnUNet owns the most model parameters. Compared to both 2D and 3D CNN-based methods, 2D transformer-based/-hybrid approaches are of lower \textit{GFLOPs} for training. Specifically, BATFormer outperforms both CNN-based and transformer-based approaches with a large margin in both \textit{GFLOPs} and \textit{Params}.
	On the ISIC 2018 dataset, distributions of approaches in the \textit{GFLOPs-Params} are more sparse, among which Attention U-Net is of the highest \textit{GFLOPs} and TransUNet contains the most model parameters. Comparatively, though the margin is not as large as that on the ACDC dataset, BATFormer is the most efficient approach with the least computational complexity.
	Based on the consistent results across the two datasets, we believe that BATFormer is much more easy-to-deploy in clinical applications.
	
	\subsection{Future Work on BATFormer}
	
	One potential way to improve BATFormer is in the window generation process of BLT, where all windows to localize boundaries share a fixed and pre-defined window size. On the one hand, the selection of the optimal window size is task-dependent as discussed in Section V.B. On the other hand, producing densely-distributed windows covering entire images for entropy calculation is less efficient. Therefore, adopting sparse windows with varying sizes would be beneficial for efficiency improvement. The main challenge is how to realize this without introducing unacceptable computational costs, which deserves more exploration in the future.
	
	\section{Conclusion}
	\label{sec:conclusion}
	
	In this paper, we propose BATFormer for lightweight and shape-preserving medical image segmentation. Specifically, a cross-scale global transformer is developed to capture multi-scale long-range dependency with lower computational complexity. Then, a boundary-aware local transformer is introduced to alleviate boundary uncertainty by adaptively generating boundary-aware windows and performing mid-range dependency establishment within each window. By jointly utilizing the transformer module and the CNN module, both global and local features are better extracted for accurate medical image segmentation.
	Experimental results on two widely-used datasets demonstrate BATFormer's superior performance against the state-of-the-art CNN-based and transformer-based approaches for medical image segmentation. Furthermore, BATFormer achieves consistent performance improvements across 2D and 3D medical image data with lower computational costs, which is feasible in clinical scenarios.

\end{document}